\documentclass[11pt,letterpaper]{article}
\usepackage[margin=0.95in]{geometry}
\usepackage[T1]{fontenc}
\usepackage[utf8]{inputenc}
\usepackage{lmodern}
\usepackage{amsmath,amssymb,booktabs,array,tabularx,graphicx}
\usepackage[numbers,sort&compress]{natbib}
\usepackage{microtype,xcolor,xurl,enumitem,tikz}
\usetikzlibrary{arrows.meta,positioning}
\usepackage{hyperref}
\usepackage[all]{hypcap}
\usepackage[section]{placeins}
\usepackage{needspace}
\hypersetup{hidelinks,pdftitle={Replayable Financial Agents: A Determinism-Faithfulness Assurance Harness for Tool-Using LLM Agents},pdfauthor={Raffi Khatchadourian},pdfsubject={A framework and empirical study of replaying financial agent decisions and tool paths}}
\setlist{leftmargin=*,itemsep=3pt,topsep=5pt}
\newcolumntype{Y}{>{\raggedright\arraybackslash}X}
\newcommand{\U}{\mathrm{U}}
\title{\textbf{Replayable Financial Agents:}\\A Determinism-Faithfulness Assurance Harness\\for Tool-Using LLM Agents}
\author{Raffi Khatchadourian\\IBM Financial Services Market, New York, NY, USA\\\small\texttt{raffi.khatchadourian1@ibm.com}}
\date{}
\begin{document}
\maketitle
\begin{abstract}
Replaying a financial agent requires specifying the conditions of a rerun,
preserving its execution record, and deciding which parts of its behavior
to compare. We introduce the Determinism-Faithfulness Assurance Harness
(DFAH), a framework and practical protocol for organizing these choices.
The framework distinguishes comparisons of final decisions, tool-name
sets, and ordered tool names, while treating support from available
evidence as a separate evaluation target. We illustrate the framework
using 2,957 recorded executions in synthetic compliance-triage and
data-operations tasks. One data-operations configuration repeated its
recorded decision across all trials on 84\% of cases, while its ordered
tool-name sequence repeated on 26\%. Conversely, some compliance cases
retained the same tool sequence while recorded decisions changed. A
case-level example shows how repeated labels accompanied different
lookups and escalation calls. These observations demonstrate how the
comparison level changes the description of repeated behavior. The
resulting protocol connects configuration control, preservation of each
attempt, comparison of records, and inspection of changed cases. DFAH
provides a practical foundation for designing replay experiments and
locating execution differences during the development and evaluation of
financial agents.
\end{abstract}
\noindent{\small\textbf{Code and data:}
\url{https://github.com/ibm-client-engineering/output-drift-financial-llms}}\par
\section{Introduction}
\label{sec:introduction}

A financial agent may investigate a transaction, look up a customer, and
decide whether to escalate an alert. Running it again can produce the same
decision through a different sequence of calls. That difference may be
harmless---for example, independent lookups occurring in a different order.
It may also change which information the agent consulted. The final label
does not distinguish these possibilities.

This paper asks how to make such behavior replayable and interpretable.
We use \emph{replay} to mean a repeated execution under a stated
configuration, followed by comparison with retained records. The
configuration includes the input, model interface, prompts, tools, and
environment. Replaying a live tool against changed data answers a different
question from rerunning the agent against a frozen tool response. An
evaluation must say which conditions it holds fixed.

The framework starts from a practical requirement: a replay result should
identify both the conditions repeated and the representation compared.
We illustrate this requirement with repeated executions in two synthetic
financial workflows. Comparing decisions, tool-name sets, and ordered
tool names reveals how the choice of comparison changes the description
of the same trials. These observations guide a protocol for designing
replay experiments and inspecting differences during development.

We introduce the Determinism-Faithfulness Assurance Harness (DFAH), a
framework for defining and comparing replays of tool-using financial
agents. It organizes the comparison into separate measurements. Decision repeatability asks whether the same
label returns. Trajectory agreement asks which parts of the tool record
repeat. Evidence-conditioned faithfulness asks whether statements in the
output are supported by the information available during that execution.
These targets use different parts of the retained execution.

This separation matters for development as well as evaluation. When a
prompt, parser, tool, or model changes, an engineer needs to locate the
resulting differences. Comparing only final answers can miss a changed
execution; comparing only tool names can miss a changed decision.
Preserving the attempts and their surrounding conditions makes those
differences available for inspection.

The paper makes three contributions:
\begin{enumerate}
\item A replay framework that distinguishes decision, tool-set, tool-sequence,
and richer record comparisons, and explains how their aggregation changes
the question being answered.
\item An empirical illustration using 12 configurations from the original
compliance and data-operations experiments. The retained records reproduce
the reported decision and tool-name agreement summaries.
\item A practical protocol connecting configuration control, execution
capture, repeated trials, and inspection of differences.
\end{enumerate}

The aim is to make replay a useful scientific and engineering operation:
state what was repeated, show what was recorded, and compare the parts
relevant to the question.

\section{Related work}
\label{sec:related}

Tool-using language agents combine generated responses with actions in an
external environment. ReAct interleaves reasoning and actions, allowing
observations from tools to influence subsequent steps \citep{yao2023react}.
AgentBench evaluates agents across interactive environments
\citep{liu2023agentbench}, while SWE-bench evaluates changes to software
repositories against tests \citep{jimenez2024swebench}. These studies
illustrate why an agent's behavior involves more than a single text output.

Repeated success is already an explicit concern in agent evaluation.
$\tau$-bench compares final database states with task goals and introduces
pass$^k$ to measure success across repeated trials \citep{yao2024taubench}.
Our focus is complementary: even when trials agree on a decision, what
parts of their recorded executions agree? Appendix~\ref{app:success}
distinguishes agreement from repeated task success.

Reproducibility research emphasizes the role of experimental conditions,
data, and evaluation procedures \citep{pineau2021checklist}. Our earlier
study of output drift examined variation across financial labeling
workflows \citep{khatchadourian2025drift}. Tool use adds another object to
compare: the sequence of calls through which an answer is produced.
Financial data operations provide a concrete setting for this question,
because investigation and remediation involve distinct tools and state
changes \citep{saini2025dataops}.

This paper introduces the DFAH framework and replay protocol. The later
DFAH-Bench study builds on this foundation to examine replay agreement,
capture loss, and authorization in additional experiments
\citep{khatchadourian2026dfah}. Its replay-faithfulness measurements concern
recorded-path fidelity; the evidence-conditioned faithfulness target
defined here concerns support for output claims. The papers share some
source records. This paper develops the replay protocol and its original
financial-task illustration; shared records are not independent replications.

\section{What a replay compares}
\label{sec:framework}

\subsection{Execution, record, and comparison level}

For an input $q$ and configuration $\theta$, an agent produces a decision
$d$ and an execution
\[
\tau=[(t_1,a_1,r_1),\ldots,(t_L,a_L,r_L)],
\]
where $t_j$ is a tool name, $a_j$ its arguments, and $r_j$ its returned
result. The length $L$ may differ across trials. A recorder retains
$R=C(\tau)$, where $C$ describes the available capture. This notation
distinguishes an execution from the information retained about it.

A comparison applies a specified representation $\phi$ to that record.
For example, $\phi_{\mathrm{set}}$ returns the set of tool names,
$\phi_{\mathrm{seq}}$ returns the ordered names including repetitions,
and $\phi_{\mathrm{args}}$ retains the ordered names and arguments.
A fuller comparison also retains results. Each answers a different
question (Table~\ref{tab:levels}).

\begin{table}[htbp]
\centering\small
\caption{Comparison levels in a replay. Richer levels require the
corresponding fields to be captured. The empirical study uses decisions,
tool-name sets, and ordered tool names.}
\label{tab:levels}
\begin{tabularx}{\linewidth}{@{}>{\raggedright\arraybackslash}p{0.23\linewidth}YY@{}}
\toprule
Level & What is compared & A difference it can reveal\\
\midrule
Decision & Final task label & A changed disposition of the same case\\
Tool-name set & Which tools appear & A lookup present in only one run\\
Tool-name sequence & Names, order, and repetitions & A repeated lookup or changed ordering\\
Names and arguments & Calls and their inputs & The same lookup directed to a different entity\\
Names, arguments, results & Calls, inputs, and returned values & A changed response to the same request\\
\bottomrule
\end{tabularx}
\end{table}

Exact equality requires a documented representation. JSON key order may
be irrelevant to a tool's input, while a changed account identifier is
usually meaningful. Canonicalization should remove representational
differences deliberately, not discard fields merely to improve an
agreement score. Missing data should remain distinguishable from an
observed empty result.

There is a simple ordering among nested record comparisons. If a coarse
representation is obtained from a richer one, agreement on the richer
representation implies agreement on the coarse one. Thus identical
ordered tool names imply identical tool-name sets; the reverse need not
hold. Decisions have no such ordering with tool-name paths: the names
alone do not determine the final label. This is why both comparisons are
needed.

\subsection{Reference agreement and case unanimity}
\label{sec:aggregation}

Suppose case $q$ has $k_q\geq2$ recorded trials. For a selected reference
trial $r$, agreement among the other trials is
\[
A_\phi(q;r)=\frac{1}{k_q-1}
\sum_{i\ne r}\mathbf{1}\{\phi(R_{qi})=\phi(R_{qr})\}.
\]
The reference's match with itself is excluded. The reference-selection
rule must be stated, since selecting the most frequent result answers a
different question from comparing with the first execution.

The experiments in this paper use \emph{case unanimity}:
\[
U_\phi(q)=\mathbf{1}\{\phi(R_{q1})=\cdots=\phi(R_{qk_q})\},\qquad
\U_\phi=\frac{1}{|Q|}\sum_{q\in Q}U_\phi(q).
\]
For decisions, replace $\phi(R_{qi})$ by $d_{qi}$. A value of 84\% means
that every recorded trial agreed on 84\% of the included cases. It does
not mean that 84\% of individual trials matched a reference.
We give each model--case group equal weight within its configuration.

Replay count affects unanimity. As a mathematical illustration, suppose
independent trials return one label with probability 0.9 and a second
with probability 0.1. The probability of unanimity is
$0.9^k+0.1^k$: 73\% for three trials and about 43\% for eight.
The system has not changed; the opportunity to observe a difference has.
Comparisons across systems should therefore match replay counts or
explicitly account for them.

``Determinism'' in the harness name refers to this measurement objective.
Finite replay agreement is an observation under specified conditions,
not a proof that all future executions will be identical.

\subsection{Evidence support}
\label{sec:evidence}

Evidence-conditioned faithfulness concerns whether an output is supported
by the evidence available in its own execution. Let $C_q$ contain the
checkable claims in an output and $E_q$ the retained evidence. Given a
specified support rule $g$, a simple claim-level summary is
\[
F_g(q)=\frac{1}{|C_q|}\sum_{c\in C_q}
\mathbf{1}\{g(c,E_q)=\text{supported}\}.
\]
This expression specifies a target, not a validated grader. An implementation
must define claim extraction, numerical tolerances, how several evidence
items may jointly support a claim, and the treatment of unresolved claims.
If no checkable claims or required evidence are available, it should report
that condition rather than assign perfect support.

For example, an output may state that an account exceeded a limit.
Assessing support requires the relevant account value, the applicable
limit, and their units and timing. Matching the same entities and numbers
does not establish that the value is above rather than below the limit.
Conversely, a paraphrase may be supported
without sharing the response's wording. This motivates preserving evidence
and calibrating any automatic support grader against task-specific
judgments.

\section{The DFAH replay protocol}
\label{sec:protocol}

The framework connects four functions: running a task, preserving its
execution, comparing records, and summarizing differences
(Figure~\ref{fig:protocol}). The protocol below specifies the design
requirements for each; Section~\ref{sec:measurements} describes the
capture available in the empirical implementation.

\begin{figure}[htbp]
\centering
\begin{tikzpicture}[
box/.style={draw=black!55,rounded corners,align=center,text width=3.0cm,
minimum height=1.12cm,font=\small,fill=black!3},
arr/.style={-{Latex[length=2mm]},thick}]
\node[box] (run) {Run a case\\Stated conditions};
\node[box,right=5mm of run] (store) {Retain a record\\Every attempt};
\node[box,right=5mm of store] (compare) {Compare\\Decisions and paths};
\node[box,right=5mm of compare] (inspect) {Inspect changes\\Report coverage};
\draw[arr] (run)--(store);
\draw[arr] (store)--(compare);
\draw[arr] (compare)--(inspect);
\end{tikzpicture}
\caption{DFAH's replay workflow. The comparison level determines which
fields the recorder must preserve. Evidence support is a separate
evaluation of an output against the evidence retained for that run.}
\label{fig:protocol}
\end{figure}

\subsection{State what is held fixed}

A replay specification identifies the case, model endpoint or local model,
prompt, tool definitions, parser, stopping rule, and environment state.
Requested sampling settings should be recorded together with the settings
actually sent to the provider. A seed is useful configuration information;
it is not an assurance that a remote service or local runtime repeats
exactly.

Two useful designs answer different questions. In a live rerun, the agent
calls tools again; variation can reflect both agent behavior and changed
responses. In a controlled replay, the experiment supplies a frozen
environment or a defined tool-response policy. A canned list of responses
is insufficient if the agent asks different questions: unmatched calls
need an explicit handling rule. The experiment should also reset simulated
state between attempts when tools can modify it.

\subsection{Retain attempts before reducing them to scores}

Each execution should have its own identifier and ordered events linking
requests to responses. Useful fields include tool names, arguments,
returned values, errors, the emitted answer, and the parser's interpretation.
Retries should remain distinguishable from the attempt they follow.
Otherwise, a later successful attempt can obscure the earlier execution.

The recorder should preserve enough information for the intended
comparison. Hashes can help detect byte changes, but cannot explain a
changed value without access to the corresponding content. A tool-name
sequence can establish that different names were recorded; it cannot
establish that requests with the same names used the same arguments.

\subsection{Compare like with like}

Define the grouping key and replay count before aggregation. A useful group
is one case under one configuration, with its repeated trials kept together.
Compare decisions, tool-name sets, and ordered names separately. Add
argument and result comparisons when capture permits them.
Record missing executions and missing fields explicitly, and report the
number of included groups alongside the rates.

An empty tool path deserves its own interpretation. Repeated empty paths
can reflect a text-only baseline, a valid no-tool decision, or missing
capture. Equality alone cannot distinguish them. The experimenter must use
the implementation and record context to say what an empty path means.

\subsection{Inspect the changed cases}

An aggregate gap is a starting point. Return to the cases with repeated
decisions and changing paths, then inspect the converse cases. A changed
order may leave the available information unchanged, whereas a changed
tool set may alter what was retrieved or which action was attempted.
The task determines whether either difference matters.

For a future intervention study, change one specified component and reuse
the same cases and replay conditions. Redeployment, stale data, missing
fields, and changed numeric inputs are useful perturbations to investigate.
A restart asks whether behavior repeats under equivalent conditions;
changing a data value asks how the agent responds to new evidence. Before
comparing conditions, specify which decisions should remain stable and
which may appropriately change. Their effects need to be measured in fresh
trials; the protocol does not
assign expected degradation rates or universal acceptance thresholds.

\section{Experiments in financial workflows}
\label{sec:experiments}

\subsection{Tasks and study design}

The study uses two synthetic tasks with 50 authored cases each.
\emph{Compliance triage} asks the agent to investigate a transaction alert
and return \texttt{escalate}, \texttt{dismiss}, or \texttt{investigate}.
Its tools supply a sanctions check, a customer profile, and a risk score.
\emph{DataOps exception handling} asks the agent to investigate a data
problem and return \texttt{auto\_fix}, \texttt{escalate}, or
\texttt{quarantine}. Its tools retrieve exception details, reference data,
and earlier fixes, and simulate validation, repair, and escalation.
The task prompts and tool interfaces are described in
Appendix~\ref{app:tasks}.

We retain 12 configurations from the original reported experiments in
these two tasks, each covering at least 25 cases. A configuration is a
model and its implemented provider path on one task. The records contain
2,957 executions in 569 model--case groups: 1,782 executions in compliance
triage and 1,175 in DataOps. There are 11 tool-calling configurations
(2,557 executions) and a Granite text-only compliance baseline
(400 executions). This is a descriptive analysis of existing records,
with no new model runs.

Local configurations used eight trials per case; hosted configurations
used three. Coverage was 50 cases except for Gemini Pro/compliance
(44) and Opus/DataOps (25). Tables~\ref{tab:compliance}
and~\ref{tab:dataops} give the denominators. The selection and public
reproduction materials are specified in Appendix~\ref{app:reproduce}.

The runner allowed up to five agent turns. Local calls requested
temperature zero and seed 42; Gemini calls requested temperature zero.
The Anthropic request omitted temperature, even though the lightweight
logs recorded zero. Its results therefore describe the provider-default
configuration. These are comparisons of implemented systems, including
their prompts, provider settings, parsers, and stopping rules.

\subsection{Recorded measurements}
\label{sec:measurements}

We compute case unanimity for the recorded decision, tool-name set, and
ordered tool-name sequence, then average cases within each configuration.
Tool-name sequences retain repetitions. A set removes both repetitions
and ordering. The implementation's earlier ``action determinism'' and
``signature determinism'' fields correspond to these set and sequence
comparisons, respectively.

The retained decisions are outputs of the runner's parser. If it could
not extract an allowed decision, the runner supplied the task's final
listed label: \texttt{investigate} for compliance or \texttt{quarantine}
for DataOps. The lightweight records do not distinguish these defaults
from labels extracted from generated text. We therefore describe
\emph{recorded decision agreement}, rather than assigning every repeated
label to a deliberate model choice.

The logs support tool-name comparisons but do not consistently preserve
arguments and results for richer replay. In particular, the runner
rewrote logs for cases with changing decisions and supplied empty
tool-output lists in that rewrite. This prevents using those records to
evaluate evidence support. The analysis uses the retained decisions and
names, without treating the tool-output hashes as complete execution
evidence.

\subsection{Decision agreement and path agreement}
\label{sec:results}

\begin{table}[htbp]
\centering\small
\caption{Compliance triage: case unanimity (\%). Rates average cases; $k$ is trials per case. Set and sequence compare recorded tool names. Granite is a text-only baseline; its empty paths are shown separately.}
\label{tab:compliance}
\begin{tabularx}{\linewidth}{@{}Yrrrrr@{}}
\toprule
Configuration & Cases & $k$ & Decision & Set & Sequence\\
\midrule
Qwen 2.5 7B & 50 & 8 & 100.0 & 100.0 & 100.0\\
GPT-OSS 20B & 50 & 8 & 86.0 & 100.0 & 92.0\\
Gemini Flash & 50 & 3 & 68.0 & 88.0 & 66.0\\
Gemini Pro & 44 & 3 & 68.2 & 59.1 & 40.9\\
Claude Sonnet 4 & 50 & 3 & 82.0 & 100.0 & 44.0\\
Claude Opus 4 & 50 & 3 & 72.0 & 98.0 & 44.0\\
\midrule
\multicolumn{6}{@{}l}{\textit{Text-only baseline (no recorded tool calls)}}\\
Granite 3.3 & 50 & 8 & 100.0 & 100.0 & 100.0\\
\bottomrule
\end{tabularx}
\end{table}

\begin{table}[htbp]
\centering\small
\caption{DataOps exception handling: case unanimity (\%). Rates average cases; $k$ is trials per case. Set and sequence compare recorded tool names. Opus covers 25 cases; the other configurations cover 50.}
\label{tab:dataops}
\begin{tabularx}{\linewidth}{@{}Yrrrrr@{}}
\toprule
Configuration & Cases & $k$ & Decision & Set & Sequence\\
\midrule
Qwen 2.5 7B & 50 & 8 & 94.0 & 94.0 & 94.0\\
GPT-OSS 20B & 50 & 8 & 84.0 & 80.0 & 74.0\\
Gemini Flash & 50 & 3 & 94.0 & 76.0 & 70.0\\
Claude Sonnet 4 & 50 & 3 & 84.0 & 56.0 & 26.0\\
Claude Opus 4 & 25 & 3 & 64.0 & 36.0 & 20.0\\
\bottomrule
\end{tabularx}
\end{table}

The compliance results show several forms of replay behavior.
Qwen repeated both its recorded decision and tool-name sequence on all
50 included cases. Sonnet repeated the decision on 41 of 50 cases
(82\%), but the sequence on 22 (44\%). Its tool-name set repeated on
all 50: the recorded variation concerned order or repetition of tools,
not which tool names appeared.

The converse also occurs. GPT-OSS repeated its tool-name sequence on
46 of 50 compliance cases (92\%), while decisions repeated on 43
(86\%). Inspection of the case groups identifies three cases with a
fixed sequence and changing decisions. A stable tool-name path therefore
does not make the recorded decision redundant.

In DataOps, Sonnet's decision unanimity was 84\% and sequence unanimity
26\%. Here the tool-name set also changed across trials on some cases:
set unanimity was 56\%. Opus showed decision and sequence unanimity
of 64\% and 20\% on its 25 included cases. These observations make the
comparison level consequential: ``the agent repeated its behavior''
would describe different fractions of cases depending on what behavior
was compared.

\begin{figure}[htbp]
\centering
\includegraphics[width=\linewidth]{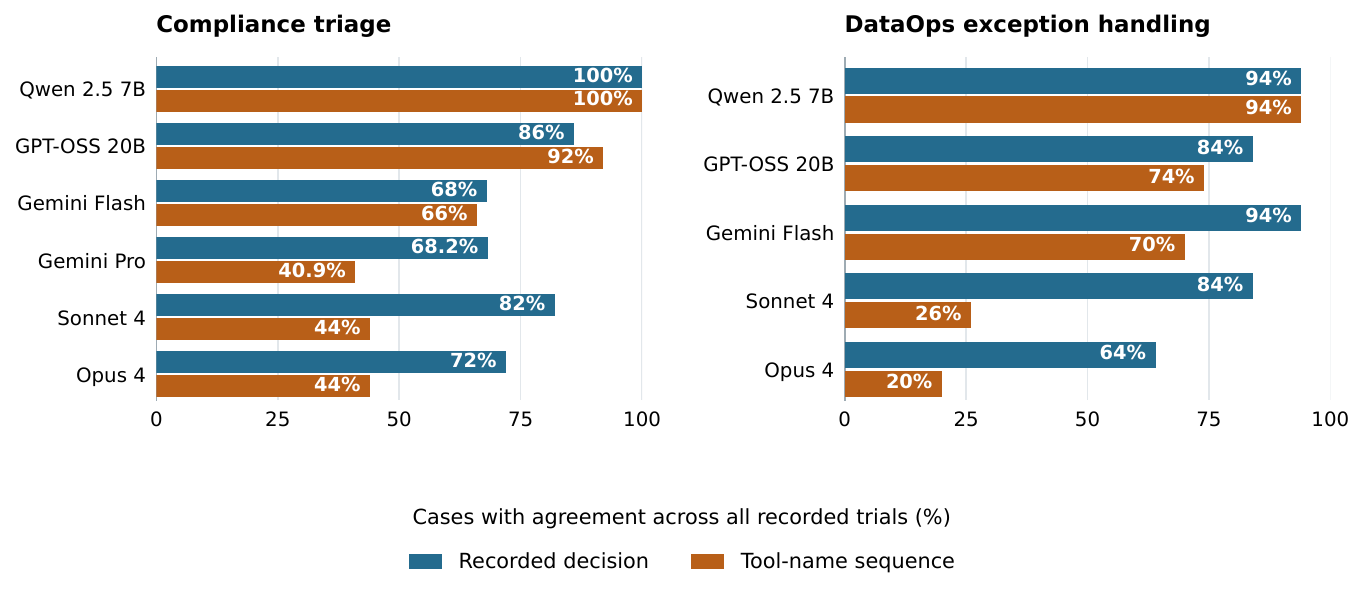}
\caption{Decision and ordered tool-name unanimity, shown separately for
each task. Paired bars compare the two measurements for each configuration;
labels give the percentage of cases with agreement across all recorded
trials. Case counts appear in Tables~\ref{tab:compliance}
and~\ref{tab:dataops}. Local configurations use eight trials per case and
hosted configurations use three. The text-only baseline appears separately
in Table~\ref{tab:compliance}.}
\label{fig:agreement}
\end{figure}

Figure~\ref{fig:agreement} displays both directions of disagreement.
The two measurements describe different parts of the same execution.
Their practical meaning becomes clearer when the changed cases are
examined directly.

\subsection{A replay viewed at the case level}
\label{sec:example}

Consider the Sonnet DataOps case DQ-001, a synthetic record containing
a negative trade price. All three recorded decisions were
\texttt{quarantine}. The first path retrieved exception details, consulted
earlier fixes twice, queried reference data, and called the simulated
escalation tool. The next two paths queried reference data earlier,
consulted earlier fixes three times, and did not record an escalation
call (Table~\ref{tab:example}).

\begin{table}[htbp]
\centering\small
\caption{Recorded Sonnet paths for DataOps case DQ-001. D = exception
details, H = historical fixes, R = reference data, and E = escalation to
human. The names abbreviate the tool calls, not their arguments or results.}
\label{tab:example}
\begin{tabularx}{\linewidth}{@{}lYl@{}}
\toprule
Trial & Ordered tool names & Recorded decision\\
\midrule
1 & D $\rightarrow$ H $\rightarrow$ H $\rightarrow$ R $\rightarrow$ E & quarantine\\
2 & D $\rightarrow$ R $\rightarrow$ H $\rightarrow$ H $\rightarrow$ H & quarantine\\
3 & D $\rightarrow$ R $\rightarrow$ H $\rightarrow$ H $\rightarrow$ H & quarantine\\
\bottomrule
\end{tabularx}
\end{table}

This case shows what inspecting a replay adds to a repeated label.
The two paths differ in both the tools present and their repetition.
The record supports those observations; it does not establish whether
an escalation completed or why the paths diverged. An execution-level
review would next inspect the arguments, responses, and simulated state.
Those are precisely the fields a richer replay needs to preserve.

\subsection{Interpreting empty and repeated paths}

The Granite compliance baseline invoked no tools in any retained
execution. Its 100\% set and sequence unanimity therefore describes
repeated empty paths. It is separated from the tool-calling configurations
because it does not demonstrate repeated use of the available tools.

Among the other configurations, one GPT-OSS DataOps group also had empty
paths in every trial. The remaining groups had at least one recorded call
in the group. We retain the empty group in that configuration's
descriptive denominator rather than silently treating it as missing.
Whether no-tool behavior is appropriate is a task question; the agreement
measurement alone cannot answer it.

Together, the results illustrate why the replay protocol specifies its
comparison level. Repeated recorded decisions can accompany changing
tool paths, and repeated tool-name paths can accompany changing decisions.
The separate measurements identify which cases to inspect and what
additional record detail would help explain the difference.

\section{Discussion}
\label{sec:discussion}

\subsection{Replay as a development tool}

The practical value of replay is the ability to investigate a change.
A team revising a financial agent can preserve the existing cases and
compare the new configuration against retained executions. If decisions
stay fixed but the path changes, the team can examine whether the
difference concerns ordering, repeated lookups, retrieved information,
or an attempted action. If the path names stay fixed but decisions
change, the team needs to inspect inputs, responses, and parsing.

Neither finding dictates that the agent must follow one approved path.
Several paths may be reasonable for the same case. The framework provides
the measurements needed to study that flexibility and to decide which
differences deserve closer examination. It also makes explicit when a
record is too coarse to answer the next question.

\subsection{From repeatability to evidence support}

Evidence support extends the replay question from ``what repeated?'' to
``what information supported this output?'' This requires preserving the
evidence available at the time of the run. A later lookup may return a
different value, and a final answer may not reveal the information used
earlier. Linking a claim to a retained response is therefore more
informative than checking that its wording repeats.

A useful next experiment would pair richer execution capture with
independently assessed claim--evidence pairs, then examine which path
changes affect support. That would test the faithfulness component of
DFAH directly. It would also help distinguish inconsequential variation
from changes that affect the basis of a decision.

\subsection{Limitations}
\label{sec:limitations}

The experiments are retrospective and descriptive. They use synthetic
cases, simulated tools, unequal case coverage, and different replay counts
across provider paths. The results characterize the included
configurations, not a population of deployed financial agents. Model
family, prompts, runtime behavior, and provider settings vary together;
the comparisons do not identify their separate causal contributions.

Interpretation also depends on the capture and parser limits described in
Section~\ref{sec:measurements}.

The study does not measure the harm or benefit of a changed path.
It also does not establish that preserving richer records improves
debugging time, reviewer decisions, or production outcomes. Those are
testable engineering questions for a prospective evaluation. Such a
study should fix the capture requirements and comparison rules before
collection and assess the consequences of observed differences.

\Needspace{16\baselineskip}
\section{Conclusion}
\label{sec:conclusion}

DFAH provides a framework and practical protocol for replaying tool-using
financial agents. It connects a stated configuration, a retained execution
record, and a defined comparison level. The compliance and DataOps
experiments illustrate how decisions, tool sets, and ordered tool names
offer different views of the same trials. Those measurements identify
cases for closer inspection and the record detail needed to investigate
them.

The resulting framework gives replay a practical purpose: locating and
understanding differences during development. Richer capture can then
connect those differences to the evidence available in a run. This
provides a foundation for studying repeatable behavior while keeping the
execution itself visible, rather than allowing its final label to stand
in for the whole process.

\section*{Author statement}
The author is responsible for the analysis and interpretation. The views
expressed are the author's and do not necessarily reflect those of his
employer.

\FloatBarrier
\clearpage
\bibliographystyle{plainnat}

\clearpage
\appendix
\section{Agreement and task success}
\label{app:success}

Agreement asks whether trials return the same result; correctness asks
whether that result satisfies a specified task goal. A constant wrong
answer has perfect agreement, while several successful executions may
follow different paths. For independent trials with fixed success
probability $p$, the probability of at least one success is
$\mathrm{pass}@k=1-(1-p)^k$; the probability that all trials succeed is
$\mathrm{pass}^k=p^k$ \citep{yao2024taubench}.
These answer different questions from the unanimity measures used here.

The original run-log field \texttt{faithfulness\_score} is a binary match
to an authored fixture label. The reproduction package retains it as
\texttt{label\_match}, averaged first within cases and then across cases.
It is neither an evidence-support measure nor part of the replay findings.

\section{Task and runner details}
\label{app:tasks}

\subsection{Compliance triage}

Each alert supplies transaction fields and a description.
The synthetic tools are:
\begin{itemize}
\item \texttt{check\_sanctions}: a sanctions-match response for the
specified entity.
\item \texttt{get\_customer\_profile}: customer risk and profile information.
\item \texttt{calculate\_risk\_score}: a numeric risk score and category.
\end{itemize}
The 50 authored cases include routine transactions and alerts involving
offshore payments, sanctions indicators, politically exposed persons,
and unusual activity. They are research fixtures, not client records or
a statement of regulatory requirements.

\subsection{DataOps exception handling}

Each exception supplies its source, field, value, rule, and record.
The synthetic tools are:
\begin{itemize}
\item \texttt{get\_exception\_details}: the exception's priority and timing.
\item \texttt{query\_reference\_data}: reference values or alternatives.
\item \texttt{get\_historical\_fixes}: candidate fixes from earlier examples.
\item \texttt{validate\_fix}: checks on a proposed repair.
\item \texttt{apply\_fix}: a simulated repair response.
\item \texttt{escalate\_to\_human}: a simulated escalation response.
\end{itemize}
The authored cases cover formatting problems, violated data rules,
reference mismatches, and missing fields. The example in
Table~\ref{tab:example} concerns a negative numeric value in the
\texttt{trade\_price} field.

\subsection{Configuration identities}

Table~\ref{tab:modelids} gives the stored model identifiers used to select
records. The identifiers describe the recorded configurations; local
model aliases alone do not identify immutable model weights.

\begin{table}[htbp]
\centering\small
\caption{Model identifiers in the released run records. Anthropic
configuration names include the endpoint version in their stored IDs.}
\label{tab:modelids}
\begin{tabularx}{\linewidth}{@{}lYl@{}}
\toprule
Display name & Stored identifier & Provider path\\
\midrule
Qwen 2.5 7B & \path{qwen2.5_7b-instruct} & Ollama\\
GPT-OSS 20B & \path{gpt-oss_20b} & Ollama\\
Granite 3.3 & \path{granite3.3_latest} & Ollama\\
Gemini Flash & \path{gemini-2.0-flash} & Gemini API\\
Gemini Pro & \path{gemini-2.5-pro} & Gemini API\\
Claude Sonnet 4 & \path{claude-sonnet-4-20250514} & Anthropic API\\
Claude Opus 4 & \path{claude-opus-4-20250514} & Anthropic API\\
\bottomrule
\end{tabularx}
\end{table}

\section{Reproducing the reported observations}
\label{app:reproduce}

The public repository is
\url{https://github.com/ibm-client-engineering/output-drift-financial-llms}.
This paper uses its records at commit
\href{https://github.com/ibm-client-engineering/output-drift-financial-llms/tree/8725fcf1741ca11f58c32468680b80a86e6b98df}{\texttt{8725fcf1741ca}}.
The paper's ancillary files provide the selection, analysis script,
derived results, and input manifest.

\subsection{Selection}

The selection retains the original seven compliance and five DataOps
configuration rows with at least 25 cases, as listed in
\path{anc/reported_configurations.csv}. It excludes the two compliance
rows with only two cases each and a Granite/DataOps configuration absent
from the original saved table.

The script reads individual JSON files from
\path{econometrics/benchmarks/results/run_logs/}, excluding duplicate
\texttt{\_full.json} files, and groups
them by model, task, and case. Run identifiers are unique within each
group. There are 319 groups with three trials and 250 with eight.

The extracted decisions and ordered tool names reproduce every retained
row of the saved table. The selected files match the author's archived
experiment snapshot byte for byte; the supplied manifest gives their paths
and hashes.

\subsection{Offline commands}

After obtaining the pinned repository and this paper's ancillary files,
run:
\begin{quote}\small
\begin{verbatim}
python3 anc/reproduce.py /path/to/repository /path/to/results
\end{verbatim}
\end{quote}
The script uses Python's standard library and Git. It writes
\path{configuration_results.csv} and \path{case_results.csv}, containing
the table summaries and their case-level indicators;
\path{worked_example.json}, containing the three records in
Table~\ref{tab:example}; and an input manifest and reproduction summary.
Copies are supplied in \path{anc/evidence/}. Recomputing from run records
requires the pinned repository. The command performs no fresh inference.

\section{Scope of the reported experiments}
\label{app:scope}

The empirical analysis uses existing compliance-triage and DataOps
execution records. It excludes the portfolio fixture because of
inconsistent totals and labels, together with the pooled correlation,
projected stress values, and unvalidated sample-size heuristic included
in earlier presentations. The retained summaries reproduce the previously
reported values, and the accompanying materials make these analyses
reproducible. No additional model runs were conducted for this revision.

\end{document}